\documentclass[10pt,twocolumn,letterpaper]{article}

\usepackage[T1]{fontenc}
\usepackage[utf8]{inputenc}
\usepackage{amsmath}
\usepackage{amsthm}                      
\usepackage{newtxtext,newtxmath}         
\usepackage[letterpaper,margin=0.78in,columnsep=0.28in]{geometry}
\usepackage{booktabs}
\usepackage{array}
\usepackage{tabularx}
\usepackage{multirow}
\usepackage{graphicx}
\usepackage[table,dvipsnames]{xcolor}
\usepackage{caption}
\usepackage{tikz}
\usetikzlibrary{arrows.meta,positioning,calc,fit,backgrounds,shapes.geometric}
\usepackage{pgfplots}
\pgfplotsset{compat=1.17}
\usepackage{microtype}
\usepackage[hidelinks]{hyperref}
\usepackage{url}
\usepackage{algorithm}
\usepackage{algpseudocode}
\algrenewcommand\algorithmicrequire{\textbf{Input:}}
\algrenewcommand\algorithmicensure{\textbf{Output:}}
\definecolor{ourshade}{gray}{0.90}
\newcommand{\best}[1]{\textbf{#1}}
\newcommand{\agentzero}{\textsc{Agent Zero}}
\newcommand{\model}[1]{\texttt{\small #1}}

\theoremstyle{plain}

\theoremstyle{definition}
\newtheorem{definition}{Definition}
\theoremstyle{remark}

\definecolor{ink}{HTML}{1C1E26}
\definecolor{subc}{HTML}{5B5F6B}
\definecolor{amberF}{HTML}{FBF3E0}\definecolor{amberS}{HTML}{C0892B}
\definecolor{grnF}{HTML}{E7F4EC}\definecolor{grnS}{HTML}{2F7D5F}
\definecolor{bluF}{HTML}{E7EEFB}\definecolor{bluS}{HTML}{3A5DA8}
\definecolor{purF}{HTML}{F0EAFA}\definecolor{purS}{HTML}{6B46C1}
\definecolor{lavF}{HTML}{F1ECFA}\definecolor{lavS}{HTML}{5B62C9}
\definecolor{rulec}{HTML}{CFCABE}
\newcommand{\dsc}{\scriptsize\color{subc}}
\newcommand{\ttl}{\bfseries\footnotesize\color{ink}}

\definecolor{mGptA}{HTML}{2A78D6}     
\definecolor{mGptSol}{HTML}{EB6834}   
\definecolor{mGptTerra}{HTML}{1BAF7A} 
\definecolor{mOpus}{HTML}{EDA100}     
\definecolor{mSonnet}{HTML}{E87BA4}   
\definecolor{mGlmFast}{HTML}{008300}  
\definecolor{mGlm}{HTML}{4A3AA7}      
\definecolor{mDeepseek}{HTML}{E34948} 
\tikzset{mdot/.style={only marks,mark=*,mark size=2.1pt,line width=0.5pt,
  mark options={draw=#1!70!black,fill=#1}}}

\usepackage{titlesec}
\titleformat{\section}{\normalfont\large\bfseries}{\thesection}{0.6em}{}
\titleformat{\subsection}{\normalfont\normalsize\bfseries}{\thesubsection}{0.6em}{}
\titlespacing*{\section}{0pt}{1.4ex plus .3ex}{0.8ex}
\titlespacing*{\subsection}{0pt}{1.1ex plus .2ex}{0.5ex}

\begin{document}

\twocolumn[{%
\begin{center}
{\LARGE\bfseries Agent Zero Memory: Provenance-Aware\\[2pt]
Long-Term Memory for LLM Agents\par}
\vspace{1.1em}
{\large
\begin{tabular}{c@{\qquad\qquad}c}
Pengyuan Zhu & Ming Wu \\[2pt]
{\small Zero Labs} & {\small Zero Labs} \\[2pt]
{\small\ttfamily pengyuan@meetzero.ai} & {\small\ttfamily ming@meetzero.ai} \\
\end{tabular}\par}
\vspace{1.4em}
\begin{minipage}{0.86\textwidth}
{\small
\textbf{Abstract.}
Large language model (LLM) agents need durable, faithful \emph{memory} of everything a user or organization has said and stored, yet most memory systems commit to a single organizing structure (a fact store, a vector index, or a knowledge graph) and inherit its blind spots. We present \emph{Agent Zero Memory}, a provenance-aware long-term memory system that distils a user's conversations, files, and connected sources into \textbf{three parallel memory systems}, each capturing a different facet of the same history: an \emph{episodic} Memory Events timeline that makes \emph{when} and \emph{what changed} first-class, an \emph{associative} entity--event knowledge graph that links people and projects across sessions, and a \emph{semantic}, curated, citation-locked \emph{Hierarchical Documentary Memory} (HDM) of durable facts. A retrieval turn runs an intent gate (so self-contained turns add no latency), a source router, and then \emph{three concurrent agentic searches}, one per system, each a tool-using loop over hybrid (embedding\,$+$\,lexical) search under agent-controlled filters; their grounded, cited answers are integrated into one answer with a single confidence. We give a formal account of the reading discipline: every learned item is a \emph{provenanced item} carrying its origin, timestamp, and evidence pointer, and every answer is read under a \emph{citation lock}, whereby it may cite only evidence its reader actually opened, so fabrication is structurally excluded and the system abstains rather than guesses. On two public benchmarks the system sets a new state of the art: \best{95.60\%} on LongMemEval and \best{93.60\%} on LoCoMo, improving over the strongest prior systems by $+0.73$ and $+1.10$ points respectively. A controlled study across eight backbone LLMs characterizes the accuracy--cost--latency frontier: accuracy varies by only $3.4$ points while per-query cost varies by $\sim\!30\times$, with near-state-of-the-art quality at up to $20\times$ lower cost per query, the signature of memory-driven, rather than model-driven, quality. \par}
\vspace{1.2em}
\end{minipage}
\end{center}
}]

\section{Introduction}
\label{sec:intro}

What a person or an organization knows is rarely written down in a usable form. It is diffused across months of conversations in which facts are established, revised, and corrected; across files and documents that state things once and never again; and across connected tools (mail, drives, wikis, code hosts) that each hold a fragment of the story. When an LLM agent is asked ``what did we decide about the launch date, and when did that change?'', answering well requires recalling a fact stated once, months ago; knowing that a later session superseded it; and citing the specific turn where each version was said. No single retrieval mechanism in today's agent stacks does all three.

\emph{Agent memory} systems~\cite{memgpt,mem0,zep} give LLMs long-term recall over conversational history, typically by extracting salient facts and storing them in a vector index or a knowledge graph. The line of work has advanced rapidly, yet it inherits a structural weakness: memory systems store \emph{what was said} but rarely model \emph{why it is believed} or how confidence in it should change as evidence accumulates; when a user corrects a fact, most systems either overwrite silently or accrete contradictions. And each commits to a \emph{single} organizing structure, a flat fact store, a vector index, or a graph, so each is strong on the question types its structure serves and weak on the rest.

We take the position that long-term memory is an exercise in maintaining a durable, queryable, and \emph{honest} model of what is known, recording not only assertions but their origin, supporting evidence, and a calibrated degree of belief, and that no single representation serves all question types. \emph{Agent Zero Memory} therefore distils a user's chats, files, and connected sources into \textbf{three interconnected memory systems}: a \emph{Memory Events} timeline that makes \emph{when} and \emph{what changed} first-class, an entity--event \emph{knowledge graph} that links people and projects across sessions, and a curated, citation-locked \emph{Hierarchical Documentary Memory} (HDM) of durable facts about the user. A retrieval turn runs an intent gate, a source router, and then \emph{three concurrent agentic searches}, one per system, whose grounded, cited answers are integrated into one. Throughout, the design follows three provenance-aware postulates (\S\ref{ssec:formal}): layered memory, citation-locked belief, and grounded, agent-controlled retrieval.

We evaluate on the two most widely used public benchmarks for long-term conversational memory: LongMemEval~\cite{longmemeval} and LoCoMo~\cite{locomo}. Our contributions are:

\begin{itemize}\itemsep2pt
\item A long-term memory architecture of three interconnected systems (a Memory Events timeline, an Entity--Event Knowledge Graph, and a curated Hierarchical Documentary Memory), built by a four-stage background pipeline and queried by three concurrent agentic searches (\S\ref{sec:method}).
\item A formal statement of the substrate's reading discipline: provenanced items and citation-locked answers (Definitions~\ref{def:prov}--\ref{def:lock}), under which fabrication is structurally excluded and the system abstains when it cannot cite (\S\ref{ssec:formal}).
\item An intent-gated, source-routed injection pipeline in which each search is a tool-using loop over hybrid embedding$+$lexical retrieval under agent-controlled filters, with citation-locked, confidence-scored answers (\S\ref{ssec:inject}).
\item State-of-the-art results on both benchmarks: 95.60\% on LongMemEval and 93.60\% on LoCoMo, exceeding the strongest reported systems by $0.73$ and $1.10$ points (\S\ref{sec:results}).
\item A controlled backbone study across eight LLMs characterising the accuracy--cost--latency trade-off, showing that a compact open model reaches 92.2\% on LongMemEval at roughly $1/20$ the per-query cost of the best configuration, together with a retrieval-channel ablation showing the embedding and lexical channels to be complementary (\S\ref{sec:results}).
\end{itemize}

\section{Related Work}
\label{sec:related}

\subsection{Long-term memory for LLM agents}
\label{ssec:related-mem}
Conversational-memory systems fall into a few families, each of which our design draws on and extends. Because the systems we compare against in \S\ref{sec:results} are the strongest reported entries on LongMemEval and LoCoMo, we analyse them here individually: how each builds and reads its memory, what that design does well, and where it stops scaling. Table~\ref{tab:related} summarizes the comparison along the dimensions that, we will argue, decide the benchmarks: memory organization, update semantics, provenance, read-cost scaling, and shareability.

\textbf{Context paging.} \textsc{MemGPT}~\cite{memgpt} treats the context window as a virtual memory hierarchy the model itself pages in and out. This removes hard context limits with no external infrastructure, but memory stays unstructured text: there is no temporal or relational organization for cross-session reasoning, and paging decisions consume model turns in the hot path of every conversation.

\textbf{Fact extraction and update.} \textsc{Mem0}~\cite{mem0} runs a two-phase pipeline: an extraction LLM distils salient candidate facts from each message pair, and a second LLM call compares each candidate against similar existing memories in a vector store and chooses an \textsc{add}/\textsc{update}/\textsc{delete} operation; later revisions add multi-signal retrieval that fuses semantic, keyword, and entity matching. The design is compact and production-friendly: per-query context stays small and latency low. But it scales poorly on both sides of the ledger. On the write side, every message pair pays LLM extraction and reconciliation calls, so ingestion cost grows linearly with history. On the epistemic side, \textsc{update} and \textsc{delete} are \emph{destructive}: the superseded value is replaced with no validity interval and no history, so the provenance of a correction, exactly what knowledge-update questions probe, is discarded; and because raw history is not the retrieval substrate, anything the extractor did not deem salient at write time is unrecoverable at read time. Notably, the authors' own evaluation found that bolting a graph onto the flat fact pool \emph{degraded} multi-hop accuracy~\cite{mem0}, evidence that relational structure added after extraction does not compose. Emergence AI's memory~\cite{emergencemem} takes the opposite bet, deliberately minimal RAG over raw turns with whole-session retrieval and a chain-of-thought reader. This is simple and fast, but with no write-time structure, contradiction resolution is deferred entirely to the reader model at query time (its weakest categories are precisely preference and knowledge-update questions), fixed retrieval hyper-parameters transfer poorly across corpora, and retrieving whole sessions makes per-query context grow with session length.

\textbf{Vector-store memory.} Supermemory~\cite{supermemory} rewrites each session into atomic memories carrying two timestamps and typed links (\emph{updates}, \emph{extends}, \emph{derives}), then answers from embedding search with only a few hundred tokens injected per query. The tiny read cost is attractive and the typed update links are a step toward versioning, but the links are themselves LLM inferences made at ingestion, with no confidence, no audit trail, and no reported precision, so a mislinked correction silently corrupts the version chain; and retrieval remains flat, so multi-hop synthesis succeeds only when every needed memory independently matches the query. Its headline figure is also a retrieval metric (recall@15) rather than end-to-end answer accuracy, which is the lower number we compare against in Table~\ref{tab:lme}. Memobase~\cite{memobase} maintains per-user \emph{profile slots} plus an event list, read by SQL lookup rather than search, which makes reads fast and predictable. The costs are structural: slots hold current values only (an update \emph{overwrites}, keeping no history), raw messages are deleted once processed (a mis-extraction can never be re-derived, and no fact can be traced to its source), the store is single-user by design with no organizational sharing, and the flat profile-plus-list shape yields its weakest results exactly on multi-hop questions.

\textbf{Observational memory.} Mastra's Observational Memory~\cite{mastra} maintains no retrieval index at all: an \emph{observer} compresses unread history into a dated observation log that lives permanently in the context window, and a \emph{reflector} rewrites the log when it exceeds a budget. The design is refreshingly infrastructure-free, its explicit dating yields strong temporal reasoning, and the stable prompt prefix caches well. But it concentrates every scaling problem in one place. Reads are query-\emph{independent}: every turn pays the full log (tens of thousands of tokens on LongMemEval) whether or not the turn needs memory, so per-turn cost scales with the lifetime of the memory rather than with the question. Capacity is capped by the context window: scaling to years of history or an organization's corpus forces ever-lossier reflection. And reflection is \emph{destructive}: superseded observations are deleted in place, leaving no audit trail from a surviving observation to its source, while each rewrite invalidates the very prompt cache the design depends on. The log is also inherently single-agent and single-user.

\textbf{Structured and curated memory.} The systems closest to ours impose explicit structure. \textsc{Zep}~\cite{zep} builds a temporal knowledge graph (Graphiti) whose bi-temporal edges are \emph{invalidated} rather than deleted on contradiction, the closest existing design to our provenance stance. Its cost is write amplification: each incoming message triggers a cascade of LLM calls (entity extraction and resolution, fact extraction, temporal parsing, invalidation), measured by an independent evaluation at over 600k tokens to ingest a single LoCoMo conversation~\cite{mem0}; its incremental community detection drifts and needs periodic full recomputation; and its own evaluation shows the graph distillation is lossy where raw context suffices. \textsc{HippoRAG}~\cite{hipporag} runs personalized PageRank over an OpenIE-extracted graph, giving single-step associative multi-hop retrieval, but quality is bottlenecked by untrained extraction, the PageRank runs over the \emph{entire} graph at query time so read cost grows with the corpus, and the graph has no temporal or update semantics: a contradicted fact simply accretes as a competing node. Hindsight~\cite{hindsight} is structurally the closest to a multi-store design, separating facts, experiences, entity summaries, and beliefs, and fusing four parallel retrievers; but every exchange passes through multi-step LLM extraction of unreported cost, and belief revision is heuristic, scalar reinforcement plus newest-wins merging, with no versioned record of what was previously believed. ByteRover~\cite{byterover} has the strongest provenance story, an LLM-curated hierarchical context tree whose every entry carries provenance and whose every operation carries a reason, but its own limitations section concedes the consequences of putting an LLM in the write path: curation cost that ``may be prohibitive'' at high ingestion rates, novel queries that fall through its cache tiers into multi-second agentic loops, an index designed for on the order of $10^4$ entries, and per-project single tenancy.

\textbf{Three recurring failure axes.} Across the families the same three scaling failures recur. \emph{(i)~Write-side:} systems that curate with an LLM per message (Mem0, Zep, Hindsight, ByteRover) pay ingestion cost linear in history, in the hot path. \emph{(ii)~Read-side:} systems that read query-independently pay for memory they do not need, whether as an ever-present log (Mastra, MemGPT) or as retrieval whose cost grows with the corpus (HippoRAG's whole-graph PageRank, Emergence's whole-session injection). \emph{(iii)~Epistemic:} systems that overwrite on update (Mem0, Memobase, Mastra's reflection, Hindsight's newest-wins merge) cannot distinguish a correction from its predecessor, and stores that are per-user or per-project by construction (Memobase, Hindsight, ByteRover) cannot become shared organizational memory.

\textbf{How we differ.} No single representation serves all question types, and no single point in the design space above escapes all three axes. Agent Zero Memory therefore maintains \emph{three} complementary memories, an \emph{episodic} events timeline (temporal / knowledge-update), an \emph{associative} entity graph (multi-hop across sessions), and a \emph{semantic}, citation-locked documentary memory (durable profile / preferences), and reads them with an \emph{intent gate}, a source router, and \emph{parallel} agentic search, so only the relevant memory is touched. Against axis~(i), distillation runs once, in the background, off the conversational hot path, and raw sources remain indexed, so nothing is unrecoverable if extraction misjudges salience. Against axis~(ii), the intent gate makes read cost query-dependent, zero for self-contained turns, and each agentic search retrieves top-$k$ from its own index rather than scanning a monolithic log or graph. Against axis~(iii), nothing is destructive: events accumulate on the timeline rather than replacing one another, the documentary memory is a deterministic mapping from curated notes, every item carries provenance, and the store is org-shareable, so a correction updates belief without erasing what was held, the property that LongMemEval~\cite{longmemeval} and LoCoMo~\cite{locomo} reward (\S\ref{ssec:why}).

\begin{table*}[t]
\centering\footnotesize
\setlength{\tabcolsep}{4.5pt}
\caption{The memory systems compared against in \S\ref{sec:results}, along the design dimensions that decide long-term-memory benchmarks. ``Read cost'' describes how per-query cost scales as the stored history grows.}
\label{tab:related}
\begin{tabular}{@{}l l l l l l@{}}
\toprule
System & Memory organization & Update semantics & Provenance & Read cost vs.\ history & Scope \\
\midrule
\rowcolor{ourshade}
\agentzero\ (ours)             & timeline $+$ graph $+$ HDM   & append, non-destructive     & citation-locked     & gated, routed, top-$k$       & org-shared \\
MemGPT~\cite{memgpt}        & paged raw context            & append (raw log)            & raw log only        & paging grows w/ history      & per agent \\
Mem0~\cite{mem0}            & flat extracted facts         & \emph{overwrite / delete}   & lost on update      & bounded (top-$k$)            & per user \\
EmergenceMem~\cite{emergencemem} & raw turns, vector index & append (raw)                & raw turns           & grows w/ session length      & per user \\
Supermemory~\cite{supermemory} & atomic memories $+$ links & LLM-inferred links          & source chunk kept   & bounded (top-$k$)            & per user \\
Memobase~\cite{memobase}    & profile slots $+$ events     & \emph{slot overwrite}       & sources deleted     & bounded (SQL)                & per user \\
Mastra~\cite{mastra}        & in-context observation log   & \emph{destructive rewrite}  & none after reflection & full log, every turn       & per agent \\
Zep~\cite{zep}              & bi-temporal entity graph     & edge invalidation           & episode links       & traversal grows w/ graph     & multi-session \\
HippoRAG~\cite{hipporag}    & OpenIE graph $+$ PPR         & accretion (no update)       & passage links       & PPR over full graph          & corpus \\
Hindsight~\cite{hindsight}  & four typed networks          & newest-wins merge           & fact$\to$source     & bounded (fused top-$k$)      & per user \\
ByteRover~\cite{byterover}  & curated context tree         & audited operations          & per-entry records   & cache cascade; LLM on miss   & per project \\
\bottomrule
\end{tabular}
\end{table*}

\subsection{Retrieval augmentation}
Retrieval-augmented generation~\cite{rag} grounds generation in retrieved passages and is the workhorse of most memory systems; dense retrievers~\cite{dpr,sbert} and the classical BM25 ranker~\cite{bm25} are complementary, and reciprocal rank fusion~\cite{rrf} combines them robustly. \textsc{GraphRAG}~\cite{graphrag} augments RAG with community summaries over an extracted entity graph. We likewise fuse dense and lexical retrieval, but each of our three agentic searches retrieves over a \emph{structured, provenanced} memory (a timeline, a typed entity--event graph, or a curated documentary memory) rather than over raw passages or an untyped entity graph, and retrieval is \emph{agent-controlled}: the reader chooses time-window, tag, source, and speaker filters and opens raw sources on demand rather than accepting a fixed top-$k$.

\section{Methodology}
\label{sec:method}

Agent Zero Memory gives an LLM agent durable recall over everything a user does: past chat sessions (including sessions with other agents), ingested files, and connected sources such as Drive, Gmail, Notion, or GitHub. Formally, it maintains a triple of stores $\mathcal{M}=(\mathcal{T},\mathcal{G}_E,\mathcal{D})$ (an events timeline, an entity--event graph, and a documentary memory), operated upon by two halves: \emph{Memory Build} constructs $\mathcal{M}$ from raw sources in the background (\S\ref{ssec:build}), and \emph{Memory Injection} retrieves, per agent turn, exactly the memory that turn requires (\S\ref{ssec:inject}).

\subsection{Problem formulation and notation}
\label{ssec:formal}
Fix a universe $\mathcal{A}$ of \emph{raw artifacts}: conversation turns, files, document chunks, and connector items. An agent's memory is a store $\mathcal{M}$ of derived items built over $\mathcal{A}$, and a query episode maps a natural-language query $q$ to a triple $(a, C, \kappa)$ comprising an answer, a citation set, and a scalar confidence $\kappa\in[0,1]$, with abstention denoted $a=\bot$.

\begin{definition}[Provenanced item]
\label{def:prov}
A \emph{provenanced item} is a pair $(x,\pi(x))$ in which $x$ is an item of the memory and $\pi(x)=\langle o(x),\, t(x),\, \mathrm{ev}(x)\rangle$ records its origin $o(x)\in\{\textsc{scan},\textsc{infer},\textsc{doc},\textsc{dialogue},\textsc{manual}\}$, an immutable creation timestamp $t(x)$, and an evidence pointer $\mathrm{ev}(x)$ resolving, possibly through intermediate summary records, to artifacts in $\mathcal{A}$.
\end{definition}

\begin{definition}[Citation-locked answer]
\label{def:lock}
Let $O\subseteq\mathcal{M}$ be the set of items a reader actually opened during an episode. An answer $(a,C,\kappa)$ is \emph{citation-locked} if $C\subseteq O$ and every atomic claim of $a$ is supported by some item of $C$. A reader unable to assemble such a $C$ must return $a=\bot$.
\end{definition}

Citation-locking is a syntactic discipline enforced by construction (the reader's interface provides no channel through which unopened material may be cited), but it carries the semantic consequence that fabrication is structurally excluded rather than merely discouraged.

\textbf{Postulates.} Three postulates recur throughout the design and, we shall argue, account for the empirical results. \textbf{(P1)~Layering.} Raw source content is segregated from the distilled memory built above it, and curated, source-locked \emph{facts} are segregated from inferred, revisable \emph{experience}; fallible inference is thereby prevented, by type discipline rather than by policy, from corrupting ground truth. \textbf{(P2)~Provenance and calibrated belief.} Every derived item is provenanced in the sense of Definition~\ref{def:prov}, carries a confidence, and is read only under the citation lock of Definition~\ref{def:lock}; the system consequently abstains rather than fabricates. \textbf{(P3)~Grounded, structure-aware retrieval.} A tool-using reader searches memory both by meaning (dense embeddings) and by exact term (lexical match) under filters it controls, and perceives data exclusively through what it retrieves; where the memory possesses navigable structure (a timeline, an entity graph, a document hierarchy), answers are produced by \emph{traversing} that structure rather than by unconstrained generation.

\subsection{Memory Build: three parallel memory systems}
\label{ssec:build}
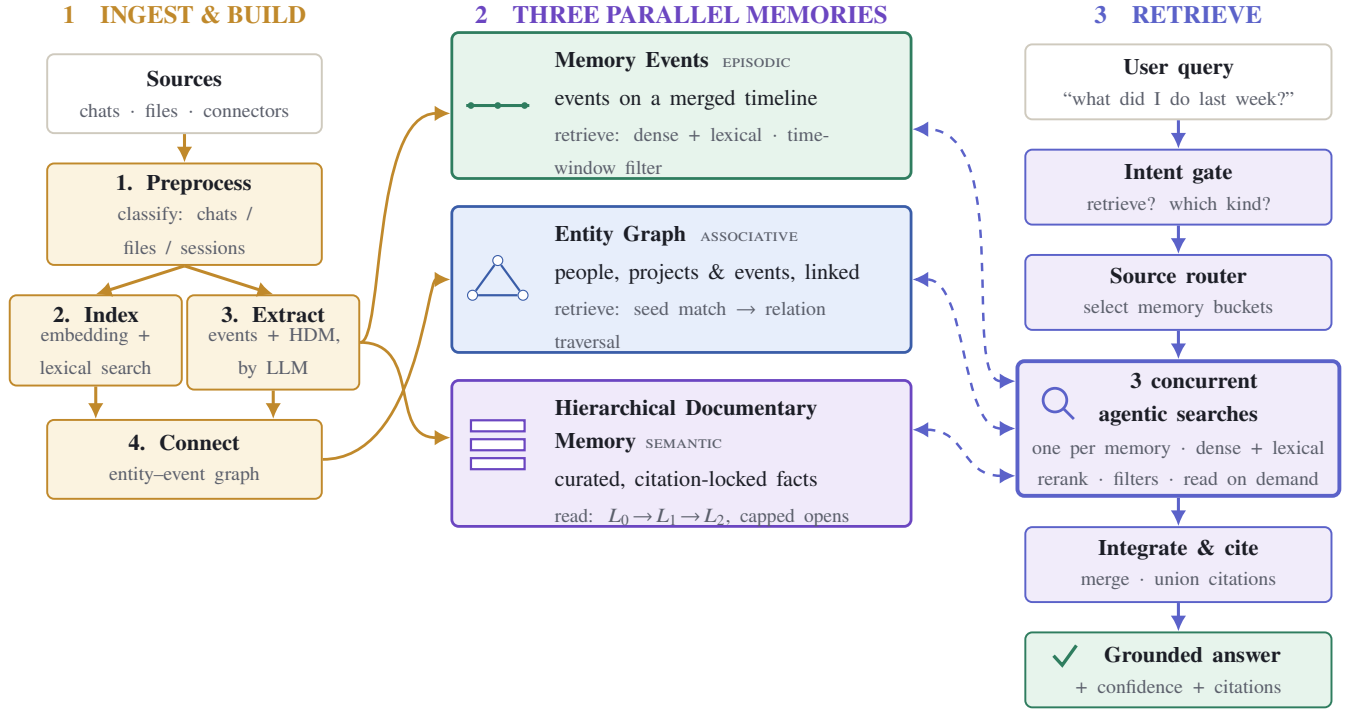
\begin{figure*}[t]
\centering
\resizebox{\textwidth}{!}{%
\begin{tikzpicture}[
  >={Latex[length=2mm,width=2mm]},font=\rmfamily,
  box/.style={rounded corners=2.5pt,align=center,inner sep=4pt,line width=0.7pt},
  stg/.style={box,draw=amberS,fill=amberF,text width=32mm,minimum height=10mm},
  store/.style={box,text width=54mm,minimum height=18.5mm,align=left,inner xsep=6pt,line width=1pt},
  qf/.style={box,draw=lavS,fill=lavF,text width=36mm,minimum height=9.5mm},
]
\node[font=\bfseries,text=amberS] at (2.0,7.9) {\small 1\quad INGEST \& BUILD};
\node[font=\bfseries,text=purS]   at (8.3,7.9) {\small 2\quad THREE PARALLEL MEMORIES};
\node[font=\bfseries,text=lavS]   at (14.6,7.9){\small 3\quad RETRIEVE};
\node[box,draw=rulec,fill=white,text width=32mm,minimum height=10mm] (src) at (2.0,6.9)
   {{\ttl Sources}\\[-1pt]{\dsc chats $\cdot$ files $\cdot$ connectors}};
\node[stg,below=3.6mm of src] (s1) {{\ttl 1. Preprocess}\\[-1pt]{\dsc classify: chats / files / sessions}};
\node[stg,text width=19mm,anchor=north] (s2) at ([xshift=-11.25mm,yshift=-3.6mm]s1.south) {{\ttl 2. Index}\\[-1pt]{\dsc embedding $+$\\ lexical search}};
\node[stg,text width=19mm,anchor=north] (s3) at ([xshift=11.25mm,yshift=-3.6mm]s1.south) {{\ttl 3. Extract}\\[-1pt]{\dsc events $+$ HDM,\\ by LLM}};
\node[stg,below=3.6mm of s3,xshift=-11.25mm] (s4) {{\ttl 4. Connect}\\[-1pt]{\dsc entity--event graph}};
\draw[->,amberS,line width=1.1pt] (src)--(s1);
\draw[->,amberS,line width=1.1pt] (s1.south)--(s2.north);
\draw[->,amberS,line width=1.1pt] (s1.south)--(s3.north);
\draw[->,amberS,line width=1.1pt] (s2.south)--(s2.south|-s4.north);
\draw[->,amberS,line width=1.1pt] (s3.south)--(s3.south|-s4.north);
\node[store,draw=grnS,fill=grnF] (me) at (8.3,6.75) {};
\node[anchor=north west,text width=40mm] at ([xshift=12mm,yshift=-4pt]me.north west)
  {{\ttl Memory Events}\, {\dsc \textsc{episodic}}\\[1.5pt]{\footnotesize\color{ink} events on a merged timeline}\\[1pt]{\dsc retrieve: dense $+$ lexical $\cdot$ time-window filter}};
\begin{scope}[shift={([xshift=6mm]me.west)}]
  \draw[grnS,line width=1.1pt] (-4mm,0)--(4mm,0);
  \foreach \x in {-3.4mm,0mm,3.4mm}{\fill[grnS](\x,0)circle(1.2pt);}
\end{scope}
\node[store,draw=bluS,fill=bluF,below=3.2mm of me] (og) {};
\node[anchor=north west,text width=40mm] at ([xshift=12mm,yshift=-4pt]og.north west)
  {{\ttl Entity Graph}\, {\dsc \textsc{associative}}\\[1.5pt]{\footnotesize\color{ink} people, projects \& events, linked}\\[1pt]{\dsc retrieve: seed match $\to$ relation traversal}};
\begin{scope}[shift={([xshift=6mm]og.west)}]
  \node[circle,draw=bluS,fill=white,inner sep=0.6pt,minimum size=3.6pt](g1)at(0,2.4mm){};
  \node[circle,draw=bluS,fill=white,inner sep=0.6pt,minimum size=3.6pt](g2)at(-3.2mm,-2mm){};
  \node[circle,draw=bluS,fill=white,inner sep=0.6pt,minimum size=3.6pt](g3)at(3.2mm,-2mm){};
  \draw[bluS,line width=0.8pt](g1)--(g2)(g1)--(g3)(g2)--(g3);
\end{scope}
\node[store,draw=purS,fill=purF,below=3.2mm of og] (wk) {};
\node[anchor=north west,text width=40mm] at ([xshift=12mm,yshift=-4pt]wk.north west)
  {{\ttl Hierarchical Documentary Memory}\ {\dsc \textsc{semantic}}\\[1.5pt]{\footnotesize\color{ink} curated, citation-locked facts}\\[1pt]{\dsc read: $L_0\!\to\!L_1\!\to\!L_2$, capped opens}};
\begin{scope}[shift={([xshift=6mm]wk.west)}]
  \draw[purS,fill=white,line width=0.8pt](-3.4mm,2.8mm)rectangle(3.4mm,4.2mm);
  \draw[purS,fill=white,line width=0.8pt](-3.4mm,0.4mm)rectangle(3.4mm,1.8mm);
  \draw[purS,fill=white,line width=0.8pt](-3.4mm,-2.0mm)rectangle(3.4mm,-0.6mm);
\end{scope}
\draw[->,amberS,line width=1pt] (s3.east) .. controls +(0:3mm) and +(180:10mm) .. ([yshift=-1mm]me.west);
\draw[->,amberS,line width=1pt] (s3.east) .. controls +(0:8mm) and +(180:8mm) .. ([yshift=2mm]wk.west);
\draw[->,amberS,line width=1pt] (s4.east) .. controls +(0:12mm) and +(180:3mm) .. (og.west);
\node[box,draw=rulec,fill=white,text width=36mm,minimum height=9.5mm] (q) at (14.6,7.05)
   {{\ttl User query}\\[-1pt]{\dsc ``what did I do last week?''}};
\node[qf,below=3.6mm of q]  (ig) {{\ttl Intent gate}\\[-1pt]{\dsc retrieve? which kind?}};
\node[qf,below=3.6mm of ig] (rt) {{\ttl Source router}\\[-1pt]{\dsc select memory buckets}};
\node[qf,below=3.6mm of rt,fill=lavF,draw=lavS,line width=1.4pt,text width=38mm,minimum height=14mm] (ps)
   {\hspace{4mm}{\ttl 3 concurrent agentic searches}\\[0pt]{\dsc one per memory $\cdot$ dense $+$ lexical}\\[-1pt]{\dsc rerank $\cdot$ filters $\cdot$ read on demand}};
\begin{scope}[shift={([xshift=5mm,yshift=3.6mm]ps.west)}]
  \draw[lavS,line width=1pt] (0,0) circle (1.5mm);
  \draw[lavS,line width=1pt] (1.1mm,-1.1mm)--(2.4mm,-2.4mm);
\end{scope}
\node[qf,below=3.6mm of ps] (in) {{\ttl Integrate \& cite}\\[-1pt]{\dsc merge $\cdot$ union citations}};
\node[box,draw=grnS,fill=grnF,text width=36mm,minimum height=9.5mm,below=3.6mm of in] (fa)
   {\hspace{3.5mm}{\ttl Grounded answer}\\[-1pt]{\dsc $+$ confidence $+$ citations}};
\begin{scope}[shift={([xshift=5mm,yshift=1.8mm]fa.west)}]
  \draw[grnS,line width=1.2pt] (-1.3mm,0)--(-0.2mm,-1.2mm)--(1.6mm,1.3mm);
\end{scope}
\draw[->,lavS,line width=1.1pt] (q)--(ig);
\draw[->,lavS,line width=1.1pt] (ig)--(rt);
\draw[->,lavS,line width=1.1pt] (rt)--(ps);
\draw[->,lavS,line width=1.1pt] (ps)--(in);
\draw[->,lavS,line width=1.1pt] (in)--(fa);
\draw[<->,lavS,dashed,line width=0.9pt] ([yshift=6mm]ps.west) .. controls +(180:7mm) and +(0:13mm) .. ([yshift=-3mm]me.east);
\draw[<->,lavS,dashed,line width=0.9pt] (ps.west) .. controls +(180:7mm) and +(0:7mm) .. (og.east);
\draw[<->,lavS,dashed,line width=0.9pt] ([yshift=-6mm]ps.west) .. controls +(180:7mm) and +(0:7mm) .. ([yshift=3mm]wk.east);
\end{tikzpicture}}
\caption{The Agent Zero Memory architecture. \textbf{(1)~Ingest \& Build:} raw sources are preprocessed and classified, then in parallel indexed (embedding\,$+$\,lexical) and distilled by an LLM into events and documentary-memory entries, which are connected into an entity--event graph. \textbf{(2)~Three parallel memories:} an \emph{episodic} Memory Events timeline (temporal / knowledge-update queries), an \emph{associative} Entity Graph (multi-hop across sessions), and a \emph{semantic}, citation-locked Hierarchical Documentary Memory (profile / preferences), each with its own retrieval mode. \textbf{(3)~Retrieve:} an intent gate decides whether a turn needs memory, a router selects buckets, then \emph{three concurrent agentic searches} (one per memory, each hybrid dense$+$lexical under agent-controlled filters) read the stores on demand, and their grounded results are integrated into one cited answer with a confidence.}
\label{fig:agentmem}
\end{figure*}

Build runs a four-stage pipeline. \emph{(1)~Preprocess \& classify} normalizes each raw source into clean text and tags it as chats, files, or agent sessions; the tag matters downstream, since chat turns, file sections, and agent traces are segmented and distilled by type-specific procedures. Stages (2) and (3) both consume this output and capture complementary views of the same material. \emph{(2)~Index} gives everything both an embedding index and a lexical-search index, so any item is findable by meaning or by lexical match, exact or fuzzy. \emph{(3)~Extract} distils the classified material, by LLM analysis, into two products: a \emph{Memory Events} timeline and a \emph{Hierarchical Documentary Memory}. \emph{(4)~Connect} runs an agentic process that builds an entity--event ontology graph over the events and documentary-memory nodes. The result is \textbf{three parallel memory systems}, each capturing a different facet of the same history (Figure~\ref{fig:agentmem}):

\textit{Memory Events}: the semantic timeline. Rather than storing only raw text, the brain distils sources into \emph{events}, each carrying a concise description of what happened and a timeline of when it happened and how it unfolded. Events are the primary unit of recall: a reader reasons over events first and opens the underlying raw chat or file only when an event needs more detail. A merged, cross-source timeline is what lets the system answer not just \emph{what} was said but \emph{when}, what changed, and which source it came from, the crux of temporal and knowledge-update questions.

\textit{Ontology Graph}: entities and relations. The agentic ``connect'' stage links the people, teams, and projects in a user's history to one another and to the events that involve them, turning a flat event stream into a navigable graph. This supports multi-hop questions whose evidence is scattered across sessions: the relevant events are reached by traversing entity relations rather than by lexical coincidence.

\textit{Hierarchical Documentary Memory (HDM)}: the curated, trustworthy core. For facts that must hold up (a user's profile, standing preferences, and the people and projects they care about), the brain keeps a separate, curated knowledge base organized into six categories (profile, preferences, entities, milestones, cases, patterns). Each entry is written at three levels of detail: an $L_0$ one-line abstract, an $L_1$ overview, and the $L_2$ full text, so a reader orients on the cheap abstract and opens the full text only when needed, under a hard cap on full opens, the \emph{hierarchical} structure the name refers to. The HDM is built by a \emph{deterministic} mapping from the user's own curated notes rather than by fact extraction, so it cannot invent or drift from what the user wrote; every entry records its source, and answers over it are \emph{citation-locked} (Definition~\ref{def:lock}) to entries actually opened, its \emph{documentary} property.

\subsection{Memory Injection: three concurrent agentic searches}
\label{ssec:inject}
Retrieval is a short pipeline of its own (Algorithm~\ref{alg:inject}, Figure~\ref{fig:agentmem}). An \emph{intent gate}, a small, fast classifier $g$, first decides whether the current turn needs memory at all, and of what kind; self-contained turns pass straight through, adding no latency. A \emph{source router} then restricts retrieval to the subset of source buckets (connectors, per-agent conversation histories, and the documentary memory) estimated most likely to contain relevant evidence. The core is three \textbf{concurrent agentic searches}, one over each memory system. Each search is itself a tool-using loop that (i)~queries its store both by meaning (embedding) and by exact term (lexical search), (ii)~controls rich filters (time window, tag, source, speaker/type) to decide autonomously what to pull, and (iii)~opens raw chunks on demand before finalizing a partial answer $(a_m, C_m, \kappa_m)$ that is citation-locked in the sense of Definition~\ref{def:lock}. A relevance rerank reorders candidates by match score, but is deliberately skipped when the reader requests a temporal order, so ``latest\,/\,since $T$'' queries stay chronological. The \emph{integrate} step then merges the three partial answers, taking $C=\bigcup_m C_m$ and an aggregated confidence, and returns a single cited answer that inherits the citation lock of its constituents.

\begin{algorithm}[t]
\small
\caption{Memory Injection (retrieval)}
\label{alg:inject}
\begin{algorithmic}[1]
\Require turn $x$; memory systems $\mathcal{M}=\{\text{Events},\text{Graph},\text{HDM}\}$
\State $g \gets \textsc{IntentGate}(x)$
\If{$\lnot\, g.\text{needs\_memory}$}
  \State \Return $\varnothing$ \Comment{no-latency pass-through}
\EndIf
\State $B \gets \textsc{Route}(x)$ \Comment{select source buckets}
\State $A \gets \textsc{parallel}\big[\,\textsc{AgenticSearch}(x,m,B)\ \text{for}\ m\in\mathcal{M}\,\big]$
\Statex \hfill\Comment{each: embedding $+$ lexical; agent-controlled filters; cited}
\State \Return $\textsc{Integrate}(A)$ \Comment{merge answers, union citations, one conf.}
\end{algorithmic}
\end{algorithm}

\subsection{Three kinds of memory, and continual learning}
\label{ssec:kinds}
Across both build and injection the brain distinguishes three kinds of memory by reliability and lifecycle (Table~\ref{tab:threekinds}): \emph{factual} memory read directly from sources (true unless the source changes); \emph{experiential} memory inferred and refined over time (verified through use, and when unclear the agent may ask); and \emph{working} memory (the current chat, instructions, and task context), which is session-bound but whose useful records are \emph{promoted} back into persistent memory when the task ends. This promotion is the continual-learning loop: what an agent learns while completing a task is summarized, archived as experiential memory related to the surrounding events, and surfaced when a later, similar task arrives. Because in an organizational deployment the memory is shared, a lesson learned by one agent becomes available to every agent.

\begin{table}[t]
\centering\small
\setlength{\tabcolsep}{4pt}
\caption{The three kinds of memory the brain draws on, by source and lifecycle.}
\label{tab:threekinds}
\begin{tabularx}{\columnwidth}{@{}l>{\raggedright\arraybackslash}X>{\raggedright\arraybackslash}X@{}}
\toprule
& Source & Lifecycle \\
\midrule
Factual & Read directly from chats / files / sources & Persistent; re-synced on change \\
Experiential & Inferred, learned from how the user works & Persistent; continuously refined \\
Working & Current chat, instructions, task context & Session-bound; useful parts promoted \\
\bottomrule
\end{tabularx}
\end{table}

\subsection{Why the design yields strong accuracy}
\label{ssec:why}
The empirical gains (\S\ref{sec:results}) follow directly from the design rather than from any single model. On the memory benchmarks, the questions that most sharply separate systems are knowledge updates, temporal reasoning, and multi-hop recall across sessions. Agent Zero Memory targets each with a dedicated system: the Memory Events \emph{timeline} makes ``when'' and ``what changed'' first-class, so a superseded fact is distinguished from its correction rather than returned stale; the entity--event \emph{graph} reaches evidence scattered across sessions by traversing relations rather than hoping for lexical overlap; and the curated, citation-locked \emph{documentary memory} answers stable profile/preference questions faithfully and, when it lacks an answer, abstains rather than guessing. Running the three searches \emph{concurrently} and integrating their cited answers gives both recall (three complementary views, each searching by meaning and by exact term) and precision (each answer is grounded and citation-locked), which is what a benchmark of knowledge updates and long-range recall rewards. Finally, because quality is governed by the structure and provenance of retrieved memory rather than a model's parametric knowledge, accuracy is stable across very different backbones (\S\ref{sec:results}), the signature we would predict from a memory-driven, rather than model-driven, design.

\section{Experimental Setup}
\label{sec:setup}

\subsection{Benchmarks and metrics}
We evaluate on the two most widely used public benchmarks for long-term conversational memory, chosen to stress precisely the capabilities our design targets.

\textbf{LongMemEval}~\cite{longmemeval} tests multi-session memory with 500 questions across six question types (single-session-user, single-session-assistant, single-session-preference, multi-session, temporal-reasoning, and knowledge-update) plus an abstention condition in which the correct behaviour is to decline when the answer was never stated. The knowledge-update and temporal types are the discriminating ones: they require distinguishing a superseded value from its correction and reasoning about \emph{when} facts held, which is exactly what our merged event timeline, with its record of when facts were asserted and superseded, is built to support. We report answer accuracy under the benchmark's LLM judge.

\textbf{LoCoMo}~\cite{locomo} evaluates very-long-term conversational memory over multi-hundred-turn, multi-session dialogues, with 1{,}540 questions across four categories: single-hop, multi-hop, temporal, and open-domain. Multi-hop and temporal questions require composing evidence retrieved from widely separated turns, stressing recall and the ability to relate facts across sessions. We again report judged answer accuracy.

For both benchmarks, Memory Build constructs the three memory systems (the events timeline, the entity--event graph, and the documentary memory) from each conversation's session history, and each held-out question is answered by the intent-gated, source-routed, three-concurrent-search injection pipeline (Algorithm~\ref{alg:inject}).

\subsection{Baselines}
We compare against the strongest publicly reported systems, including graph- and vector-based memories (\textsc{Zep}~\cite{zep}, \textsc{Mem0}~\cite{mem0}, Mastra, Hindsight, EmergenceMem, Supermemory, ByteRover, Memobase). All competitor numbers are the best publicly reported figures for each system.

\subsection{Backbones}
Agent Zero Memory is backbone-agnostic: the memory stores, indexes, retriever, and control logic are fixed, and only the LLM that drives inference and reasoning is varied. Unless noted, headline results use the best configuration per benchmark (\S\ref{sec:results}). We additionally run a controlled study over eight backbones spanning three vendors and a wide capability/price range: \model{gpt-5.5}, \model{gpt-5.6-sol}, \model{gpt-5.6-terra}, \model{gpt-5.4-mini}, \model{opus4.8}, \model{sonnet5}, \model{glm5.2}, \model{glm5.2fast}, and \model{deepseek-v4-pro}.

\subsection{Implementation and measurement}
The memory stores and both indexes are backed by a single PostgreSQL instance: events, entity--event graph nodes and edges, and documentary-memory entries are stored relationally with their provenance, dense retrieval uses \texttt{pgvector} over 3072-dimensional embeddings under cosine similarity, and lexical search combines BM25 ranking~\cite{bm25} with fuzzy (approximate) string matching for spelling and morphological variants. Each agentic search draws the top-$K$ from each channel and fuses them by reciprocal rank fusion~\cite{rrf} at $k=60$ before the optional relevance rerank (\S\ref{ssec:inject}). We report three quantities per query: median wall-clock latency (robust to a small number of slow tails), average end-to-end monetary cost computed from each provider's public list prices for the exact prompt / cached-input / completion token counts we log, and those token counts themselves. Cached-input tokens are billed at the provider's reduced cached rate, which is why token-heavy backbones are not always the most expensive. All accuracies are single-run under the benchmark-specified protocol with the official LLM judge.

\section{Results}
\label{sec:results}

\subsection{Main results}
Tables~\ref{tab:lme} and~\ref{tab:locomo} report Agent Zero Memory against the strongest reported systems on LongMemEval and LoCoMo. On LongMemEval we reach \best{95.60\%}, ahead of the next-best system (Mastra, 94.87\%) by $0.73$ points and well clear of graph- and vector-memory baselines such as Zep (71.20\%). On LoCoMo we reach \best{93.60\%}, ahead of \textsc{Mem0} (92.50\%) and ByteRover~2.0 (92.20\%). These margins set a new state of the art on both mature, near-saturated benchmarks, and, as the backbone study below shows, are achieved by several distinct backbones, indicating that the gains come from the memory substrate rather than from a single model.

\begin{table}[t]
\centering\small
\caption{LongMemEval (500 questions, 6 categories). Judged answer accuracy.}
\label{tab:lme}
\begin{tabularx}{\columnwidth}{@{}c>{\raggedright\arraybackslash}X l r@{}}
\toprule
\# & System & Note & Acc.\ (\%) \\
\midrule
\rowcolor{ourshade}
1 & \best{\agentzero} & New 2026 algorithm & \best{95.60} \\
2 & Mastra & Observational memory, 2026 & 94.87 \\
3 & Hindsight & Vectorize, 2025 & 91.40 \\
4 & EmergenceMem & Emergence AI, 2026 & 86.00 \\
5 & Supermemory & Supermemory, 2026 & 85.20 \\
6 & Zep~\cite{zep} & Open-source memory layer & 71.20 \\
\bottomrule
\end{tabularx}
\end{table}

\begin{table}[t]
\centering\small
\caption{LoCoMo (1{,}540 questions, 4 categories). Judged answer accuracy.}
\label{tab:locomo}
\begin{tabularx}{\columnwidth}{@{}c>{\raggedright\arraybackslash}X l r@{}}
\toprule
\# & System & Note & Acc.\ (\%) \\
\midrule
\rowcolor{ourshade}
1 & \best{\agentzero} & New 2026 algorithm & \best{93.60} \\
2 & Mem0~\cite{mem0} & Mem0 research, 2026 & 92.50 \\
3 & ByteRover 2.0 & Agent memory, 2026 & 92.20 \\
4 & Hindsight & Vectorize, 2025 & 89.60 \\
5 & Memobase & Open-source memory & 75.80 \\
6 & Zep~\cite{zep} & Open-source memory layer & 75.10 \\
\bottomrule
\end{tabularx}
\end{table}

\subsection{Accuracy, cost, and latency across backbones}
\begin{figure}[t]
\centering
\begin{tikzpicture}
\begin{axis}[
  width=\columnwidth,height=5.2cm,
  xmode=log,log basis x=10,
  xlabel={Avg.\ cost per query (USD, log scale)},
  ylabel={Accuracy (\%)},
  xlabel style={font=\footnotesize},ylabel style={font=\footnotesize},
  tick label style={font=\scriptsize},
  ymin=91.4,ymax=96.4,
  xmin=0.0013,xmax=0.09,
  grid=both,grid style={gray!18},
]
\addplot[mdot=mGptA]      coordinates {(0.034788,95.60)};
\addplot[mdot=mGptSol]    coordinates {(0.030663,95.20)};
\addplot[mdot=mGptTerra]  coordinates {(0.018913,94.00)};
\addplot[mdot=mOpus]      coordinates {(0.052221,93.80)};
\addplot[mdot=mSonnet]    coordinates {(0.028520,93.20)};
\addplot[mdot=mGlmFast]   coordinates {(0.014139,93.00)};
\addplot[mdot=mGlm]       coordinates {(0.010525,92.60)};
\addplot[mdot=mDeepseek]  coordinates {(0.001768,92.20)};
\node[font=\scriptsize,anchor=south east] at (axis cs:0.034788,95.60) {\model{gpt-5.5}};
\node[font=\scriptsize,anchor=east]  at (axis cs:0.030663,95.20) {\model{gpt-5.6-sol}\,};
\node[font=\scriptsize,anchor=east]  at (axis cs:0.018913,94.00) {\model{gpt-5.6-terra}\,};
\node[font=\scriptsize,anchor=east]  at (axis cs:0.052221,93.80) {\model{opus4.8}\,};
\node[font=\scriptsize,anchor=west]  at (axis cs:0.028520,93.20) {\,\model{sonnet5}};
\node[font=\scriptsize,anchor=east]  at (axis cs:0.014139,93.00) {\model{glm5.2fast}\,};
\node[font=\scriptsize,anchor=west]  at (axis cs:0.010525,92.60) {\,\model{glm5.2}};
\node[font=\scriptsize,anchor=west]  at (axis cs:0.001768,92.20) {\,\model{deepseek-v4-pro}};
\end{axis}
\end{tikzpicture}
\caption{Accuracy--cost frontier on LongMemEval across eight backbones (data from Table~\ref{tab:lme-internal}). Accuracy varies by only $3.4$ points while per-query cost varies by $\sim\!30\times$; several backbones are Pareto-efficient. Each backbone keeps the same color here and in Figure~\ref{fig:latency}.}
\label{fig:pareto}
\end{figure}
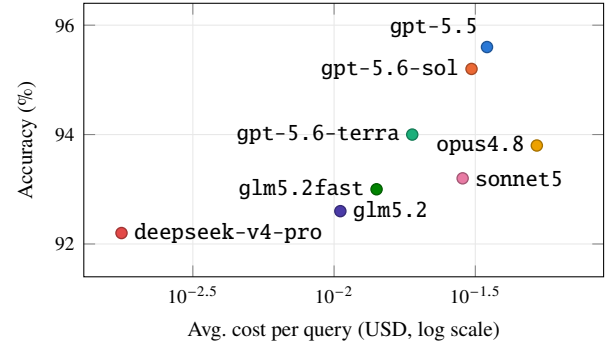

Because the substrate is backbone-agnostic, the practical question is not only how high accuracy can go but at what cost and latency. Table~\ref{tab:lme-internal} varies only the pipeline's main model on LongMemEval. Accuracy is remarkably stable across a wide range of backbones (all eight lie within $3.4$ points, from 92.20\% to 95.60\%), confirming that the memory substrate, not the model, drives quality. The models differ sharply, however, in cost and speed. The best-accuracy configuration (\model{gpt-5.5}, 95.60\%) is neither the fastest nor the cheapest; \model{glm5.2fast} answers a question in a median $6.16$\,s (fastest) at 93.00\%, while \model{deepseek-v4-pro} reaches 92.20\% at \$0.001768 per query, roughly $20\times$ cheaper than \model{gpt-5.5} for a $3.4$-point accuracy reduction. Figure~\ref{fig:pareto} plots this frontier: several backbones are Pareto-efficient, giving deployers a principled choice between top accuracy and low cost. Table~\ref{tab:locomo-internal} reports the corresponding study on LoCoMo; the same pattern holds, with \model{sonnet5} best (93.57\%).

\begin{table*}[t]
\centering\small
\caption{Backbone study on \textbf{LongMemEval}. Only the pipeline's main model is varied; memory and retrieval configuration are held fixed. Cost is average end-to-end USD per query from provider list prices; latency is per-query median.}
\label{tab:lme-internal}
\begin{tabular}{l r r r r r r}
\toprule
Backbone & Accuracy (\%) & Median latency (s) & Avg.\ cost (USD) & Avg.\ prompt tok. & Avg.\ cached-in tok. & Avg.\ compl.\ tok. \\
\midrule
\rowcolor{ourshade}
\model{gpt-5.5}         & \best{95.60} & 16.311 & 0.034788 & 12{,}224.8 & 5{,}324.8  & 446.1 \\
\model{gpt-5.6-sol}     & 95.20 & 15.780 & 0.030663 & 11{,}234.8 & 5{,}990.4  & 406.5 \\
\model{gpt-5.6-terra}   & 94.00 & 21.183 & 0.018913 & 11{,}881.0 & 6{,}412.8  & 390.6 \\
\model{opus4.8}         & 93.80 & 12.623 & 0.052221 & 23{,}779.1 & 14{,}897.8 & 591.0 \\
\model{sonnet5}         & 93.20 & 10.754 & 0.028520 & 24{,}520.4 & 17{,}807.1 & 592.1 \\
\model{glm5.2fast}      & 93.00 & \best{6.158} & 0.014139 & 10{,}986.4 & 5{,}118.0 & 339.1 \\
\model{glm5.2}          & 92.60 & 8.314 & 0.010525 & 10{,}986.1 & 4{,}630.1 & 341.4 \\
\model{deepseek-v4-pro} & 92.20 & 16.877 & \best{0.001768} & 17{,}436.1 & 15{,}372.8 & 937.0 \\
\bottomrule
\end{tabular}
\end{table*}

\begin{table*}[t]
\centering\small
\caption{Backbone study on \textbf{LoCoMo}. Conventions as in Table~\ref{tab:lme-internal}.}
\label{tab:locomo-internal}
\begin{tabular}{l r r r r r r}
\toprule
Backbone & Accuracy (\%) & Median latency (s) & Avg.\ cost (USD) & Avg.\ prompt tok. & Avg.\ cached-in tok. & Avg.\ compl.\ tok. \\
\midrule
\rowcolor{ourshade}
\model{sonnet5}         & \best{93.57} & 22.855 & 0.047083 & 45{,}818.7 & 35{,}987.8 & 1{,}279.8 \\
\model{opus4.8}         & 91.49 & 23.530 & 0.056355 & 33{,}816.2 & 26{,}466.6 & 1{,}102.2 \\
\model{gpt-5.6-sol}     & 91.49 & 18.259 & 0.039919 & 14{,}791.3 & 7{,}288.3  & 685.0 \\
\model{deepseek-v4-pro} & 90.45 & 34.741 & 0.010733 & 49{,}057.8 & 38{,}247.7 & 2{,}533.1 \\
\model{glm5.2fast}      & 90.45 & 17.042 & 0.022719 & 31{,}340.9 & 22{,}650.3 & 1{,}021.6 \\
\model{gpt-5.6-terra}   & 88.38 & 18.353 & 0.023848 & 16{,}778.7 & 9{,}162.2  & 751.4 \\
\model{gpt-5.4-mini}    & 81.10 & 15.959 & \best{0.009004} & 13{,}817.6 & 7{,}344.6 & 799.6 \\
\bottomrule
\end{tabular}
\end{table*}

\subsection{Operating points and cost efficiency}
\label{ssec:efficiency}
\begin{figure}[t]
\centering
\begin{tikzpicture}
\begin{axis}[
  width=\columnwidth,height=5.0cm,
  xlabel={Median latency per query (s)},
  ylabel={Accuracy (\%)},
  xlabel style={font=\footnotesize},ylabel style={font=\footnotesize},
  tick label style={font=\scriptsize},
  ymin=91.4,ymax=96.4,
  xmin=4,xmax=24,
  grid=both,grid style={gray!18},
]
\addplot[mdot=mGptA]      coordinates {(16.311,95.60)};
\addplot[mdot=mGptSol]    coordinates {(15.780,95.20)};
\addplot[mdot=mGptTerra]  coordinates {(21.183,94.00)};
\addplot[mdot=mOpus]      coordinates {(12.623,93.80)};
\addplot[mdot=mSonnet]    coordinates {(10.754,93.20)};
\addplot[mdot=mGlmFast]   coordinates {(6.158,93.00)};
\addplot[mdot=mGlm]       coordinates {(8.314,92.60)};
\addplot[mdot=mDeepseek]  coordinates {(16.877,92.20)};
\node[font=\scriptsize,anchor=south] at (axis cs:16.311,95.60) {\model{gpt-5.5}};
\node[font=\scriptsize,anchor=east]  at (axis cs:15.780,95.20) {\model{gpt-5.6-sol}\,};
\node[font=\scriptsize,anchor=east]  at (axis cs:21.183,94.00) {\model{gpt-5.6-terra}\,};
\node[font=\scriptsize,anchor=east]  at (axis cs:12.623,93.80) {\model{opus4.8}\,};
\node[font=\scriptsize,anchor=south west] at (axis cs:10.754,93.20) {\,\model{sonnet5}};
\node[font=\scriptsize,anchor=west]  at (axis cs:6.158,93.00) {\,\model{glm5.2fast}};
\node[font=\scriptsize,anchor=west]  at (axis cs:8.314,92.60) {\,\model{glm5.2}};
\node[font=\scriptsize,anchor=north] at (axis cs:16.877,92.20) {\model{deepseek-v4-pro}};
\end{axis}
\end{tikzpicture}
\caption{Accuracy--latency frontier on LongMemEval (data from Table~\ref{tab:lme-internal}), complementary to the accuracy--cost view of Figure~\ref{fig:pareto}; point colors identify the same backbones in both figures. The fastest backbone (\model{glm5.2fast}, $6.16$\,s) is within $2.6$ points of the most accurate one.}
\label{fig:latency}
\end{figure}
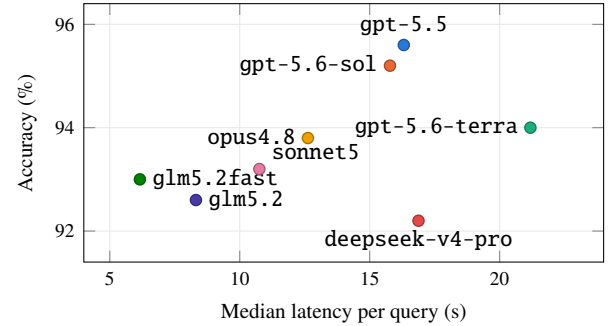

The stability of accuracy across backbones turns model choice into an economic decision, and the substrate makes that decision explicit. Two views of the same eight configurations are informative. Figure~\ref{fig:pareto} plots accuracy against cost; Figure~\ref{fig:latency} plots accuracy against latency, exposing a second, independent frontier. No single backbone dominates all three axes: \model{gpt-5.5} maximizes accuracy, \model{glm5.2fast} minimizes latency ($6.16$\,s), and \model{deepseek-v4-pro} minimizes cost (\$0.001768). To quantify the trade-off we consider cost efficiency, the accuracy delivered per unit spend. Normalizing to accuracy points per US cent, the eight LongMemEval configurations span more than an order of magnitude, from $18.0$ (\model{opus4.8}) and $27.5$ (\model{gpt-5.5}) to $88.0$ (\model{glm5.2}) and $521$ (\model{deepseek-v4-pro}), so the most cost-efficient backbone delivers roughly $19\times$ more accuracy per dollar than the most accurate one, while giving up only $3.4$ points. In practice this lets an operator pick an operating point by policy: latency-critical interactive use favors \model{glm5.2fast}; high-volume batch use favors \model{deepseek-v4-pro} or \model{glm5.2}; accuracy-critical use favors \model{gpt-5.5} or \model{gpt-5.6-sol}. Because the substrate is shared and backbone-agnostic, these points can even be mixed within one deployment without rebuilding the memory.

\textbf{Where the cost goes.} Cost is not a simple function of raw token volume, because providers bill cached-input tokens at a reduced rate and our retrieval-heavy prompts are highly cacheable. On LongMemEval (Table~\ref{tab:lme-internal}), the cached fraction of the input (cached-input over prompt tokens) ranges from $43.6\%$ (\model{gpt-5.5}) to $88.2\%$ (\model{deepseek-v4-pro}); \model{opus4.8} and \model{sonnet5} process the most tokens ($23.8$k and $24.5$k prompt tokens) yet are not the most expensive, precisely because a large share is cached. \model{deepseek-v4-pro}'s combination of a low base price and an $88\%$ cache rate is what makes it the cost leader despite emitting the most completion tokens. This is actionable: the substrate's stable, reusable context is what makes aggressive prompt caching effective, so a deployer optimizing cost should prefer backbones with both a low base price and a favorable cached-token discount.

\subsection{Ablation: hybrid retrieval}
\label{ssec:ablation}
Each agentic search queries its store through two channels, dense embeddings and lexical match (\S\ref{ssec:inject}); to quantify their joint contribution we ablate the retrieval interface. Holding the backbone fixed at \model{gpt-5.6-sol} and leaving every other component of the pipeline unchanged, we expose to the agentic searches a single restricted retrieval channel and re-run LongMemEval (Table~\ref{tab:ablation}): the embedding channel alone, the full lexical channel alone (BM25 ranking with fuzzy matching, \S\ref{sec:setup}), or a deliberately minimal exact-substring (grep-style) matcher with neither ranking nor fuzzy matching. Restricting retrieval to embeddings alone costs $1.2$ points ($95.20\%\!\to\!94.00\%$); the lexical-only and grep-only variants cost $1.8$ and $1.6$ points respectively. Two observations follow. First, the channels are genuinely complementary: no single channel recovers the hybrid's accuracy, consistent with the standard division of labour in which embeddings retrieve paraphrased or semantically related evidence with no lexical overlap, while lexical match retrieves exact identifiers, names, and rare terms that embedding similarity blurs. Second, every restricted variant still exceeds the strongest external baseline in Table~\ref{tab:lme}, indicating that the structured, provenanced stores and the agentic reading discipline, rather than any one retrieval channel, carry most of the quality; the hybrid retriever contributes the final margin that sets the state of the art.

\begin{table}[t]
\centering\small
\caption{Retrieval-channel ablation on LongMemEval (backbone \model{gpt-5.6-sol}; all other components fixed). The agentic searches are restricted to a single retrieval channel.}
\label{tab:ablation}
\begin{tabularx}{\columnwidth}{@{}>{\raggedright\arraybackslash}X r r@{}}
\toprule
Retrieval exposed to agent & Acc.\ (\%) & $\Delta$ \\
\midrule
\rowcolor{ourshade}
embedding $+$ lexical  & \best{95.20} & --- \\
Embedding & 94.00 & $-1.20$ \\
grep & 93.60 & $-1.60$ \\
Lexical & 93.40 & $-1.80$ \\
\bottomrule
\end{tabularx}
\end{table}

\section{Discussion}
\label{sec:discussion}

The design treats long-term memory as the maintenance of an honest model of what is known: a knowledge update is a new, timestamped belief recorded alongside its evidence rather than a destructive overwrite, so the correct, most-recent fact is always recoverable with its provenance, and the superseded fact remains distinguishable from its correction. The specialization across the three memories matters (the timeline is tuned to temporal and knowledge-update questions, the entity graph to multi-hop questions, and the documentary memory to durable profile and preference facts), but the shared layered-memory and citation-locked-provenance principles are what make all three faithful.

\textbf{Quality is retrieval-dominated.} In the memory setting, quality is dominated by \emph{retrieval and representation}: once the correct, provenance-tagged evidence is surfaced, even compact backbones answer correctly, which is why the LongMemEval spread across eight very different backbones is only $3.4$ points and the state of the art is reached by more than one model family. This is the signature we would expect if quality is governed by the structure and provenance of the retrieved memory rather than by any one model's parametric knowledge, and it is a useful design signal: investment in retrieval and provenance pays off across all backbones.

\textbf{Deployment economics.} Because near-top accuracy is available from compact backbones, deployers can trade a small amount of accuracy for large savings: on LongMemEval, \model{deepseek-v4-pro} is within $3.4$ points of the best system at $\sim\!1/20$ the cost, and \model{glm5.2fast} is the fastest at $6.16$\,s median latency (\S\ref{ssec:efficiency}).

\textbf{Practical deployment.} Two properties matter beyond benchmark scores. First, because every answer carries source references down to a conversation turn or file, outputs are auditable: an operator can inspect \emph{why} the system believes what it reported, which is a prerequisite for enterprise adoption. Second, because Memory Build runs in the background and the stores are shared and backbone-agnostic, the memory can be built incrementally over large sources and served by different backbones at different operating points without rebuilding, the flexibility quantified in \S\ref{ssec:efficiency}.

\textbf{Threats to validity.} Competitor figures are the best publicly reported numbers, each obtained under its own harness; differences in judge model, prompting, and retrieval budget mean the cross-system comparisons (Tables~\ref{tab:lme}--\ref{tab:locomo}) should be read as situating our system among the strongest reported results rather than as perfectly controlled head-to-head trials. The backbone study (Tables~\ref{tab:lme-internal}--\ref{tab:locomo-internal}), by contrast, \emph{is} controlled: identical memory, retriever, and control logic, varying only the model, so its internal comparisons are apples-to-apples. Costs are computed from list prices at evaluation time and will drift as prices change; we therefore report token counts alongside cost so the analysis can be repriced.

\textbf{Limitations.} Our evaluation follows the benchmarks' LLM-judged protocol. The backbone identifiers reflect the model lineup available at evaluation time. Beyond the retrieval-channel ablation (\S\ref{ssec:ablation}), a natural next step is a component-level ablation of the three memories and the intent gate; \S\ref{ssec:why} already ties each component to the benchmark failure modes it addresses, and quantifying each mechanism's marginal contribution is planned future work.

\section{Conclusion}
\label{sec:conclusion}
We presented \emph{Agent Zero Memory}, a provenance-aware long-term memory system that distils a user's conversations, files, and connected sources into three parallel memory systems (an episodic events timeline, an associative entity--event graph, and a semantic, citation-locked documentary memory) and answers a query with three concurrent agentic searches whose grounded, cited results are integrated into one answer. It sets a new state of the art on LongMemEval (95.60\%) and LoCoMo (93.60\%), and does so robustly across eight backbone models, offering a favorable accuracy--cost--latency frontier for practical deployment. We view the provenance-aware, layered memory substrate (rather than any single model or any per-query pipeline) as the durable abstraction for long-term agent memory.

Several directions follow naturally. \emph{(i)~Component ablations:} extending the retrieval-channel ablation of \S\ref{ssec:ablation} to matched-number studies that remove one of the three memories or the intent gate, quantifying each mechanism's marginal contribution to the results reported here. \emph{(ii)~Calibration:} turning reported answer confidences into probabilities with a frequentist reading, so that a stated confidence of $0.9$ is empirically justified. \emph{(iii)~Conflict arbitration:} principled selection when several valid versions of a fact coexist across sources, including when to ask the user rather than choose. \emph{(iv)~Forgetting:} graceful decay of stale experiential memory: down-weighting rather than deletion, with tombstones that preserve auditability. We regard the honest, auditable modelling of \emph{what a user or organization knows and why it believes it} as the central problem, and the layered, provenance-aware memory as a durable step toward it.



\begin{thebibliography}{99}
\small
\bibitem{longmemeval} D.~Wu, H.~Wang, W.~Yu, Y.~Zhang, K.-W.~Chang, and D.~Yu.
LongMemEval: Benchmarking Chat Assistants on Long-Term Interactive Memory.
\emph{ICLR}, 2025. arXiv:2410.10813.

\bibitem{locomo} A.~Maharana, D.-H.~Lee, S.~Tulyakov, M.~Bansal, F.~Barbieri, and Y.~Fang.
Evaluating Very Long-Term Conversational Memory of LLM Agents.
\emph{ACL}, 2024. arXiv:2402.17753.

\bibitem{memgpt} C.~Packer, S.~Wooders, K.~Lin, et al.
MemGPT: Towards LLMs as Operating Systems.
arXiv:2310.08560, 2023.

\bibitem{mem0} P.~Chhikara, D.~Khant, S.~Aryan, T.~Singh, and D.~Yadav.
Mem0: Building Production-Ready AI Agents with Scalable Long-Term Memory.
arXiv:2504.19413, 2025.

\bibitem{zep} P.~Rasmussen, P.~Paliychuk, T.~Beauvais, J.~Ryan, and D.~Chalef.
Zep: A Temporal Knowledge Graph Architecture for Agent Memory.
arXiv:2501.13956, 2025.

\bibitem{hipporag} B.~J.~Guti\'errez, Y.~Shu, Y.~Gu, M.~Yasunaga, and Y.~Su.
HippoRAG: Neurobiologically Inspired Long-Term Memory for Large Language Models.
\emph{NeurIPS}, 2024. arXiv:2405.14831.

\bibitem{graphrag} D.~Edge, H.~Trinh, N.~Cheng, et al.
From Local to Global: A Graph RAG Approach to Query-Focused Summarization.
arXiv:2404.16130, 2024.

\bibitem{rag} P.~Lewis, E.~Perez, A.~Piktus, et al.
Retrieval-Augmented Generation for Knowledge-Intensive NLP Tasks.
\emph{NeurIPS}, 2020.

\bibitem{dpr} V.~Karpukhin, B.~O\u{g}uz, S.~Min, et al.
Dense Passage Retrieval for Open-Domain Question Answering.
\emph{EMNLP}, 2020.

\bibitem{sbert} N.~Reimers and I.~Gurevych.
Sentence-BERT: Sentence Embeddings using Siamese BERT-Networks.
\emph{EMNLP}, 2019.

\bibitem{bm25} S.~Robertson and H.~Zaragoza.
The Probabilistic Relevance Framework: BM25 and Beyond.
\emph{Foundations and Trends in Information Retrieval}, 3(4):333--389, 2009.

\bibitem{rrf} G.~V.~Cormack, C.~L.~A.~Clarke, and S.~B\"uttcher.
Reciprocal Rank Fusion Outperforms Condorcet and Individual Rank Learning Methods.
\emph{SIGIR}, 2009.

\bibitem{hindsight} C.~Latimer, N.~Boschi, A.~Neeser, C.~Bartholomew, G.~Srivastava, X.~Wang, and N.~Ramakrishnan.
Hindsight is 20/20: Building Agent Memory that Retains, Recalls, and Reflects.
arXiv:2512.12818, 2025.

\bibitem{byterover} A.~Nguyen, D.~Doan, H.~Pham, et al.
ByteRover: Agent-Native Memory Through LLM-Curated Hierarchical Context.
arXiv:2604.01599, 2026.

\bibitem{emergencemem} Emergence AI.
SOTA on LongMemEval with RAG. Technical report, 2026.
\url{https://www.emergence.ai/blog/sota-on-longmemeval-with-rag}.

\bibitem{supermemory} Supermemory.
LongMemEval. Research report, 2026.
\url{https://supermemory.ai/research/longmembench/}.

\bibitem{mastra} Mastra.
Observational Memory: 95\% on LongMemEval. Research report, 2026.
\url{https://mastra.ai/research/observational-memory}.

\bibitem{memobase} Memobase.
Memobase: profile-based long-term user memory for AI. Open-source system, 2025.
\url{https://github.com/memodb-io/memobase}.

\end{thebibliography}
\end{document}